\def\year{2016}
\documentclass[letterpaper]{article}
\usepackage{aaai16}
\usepackage{times}
\usepackage{helvet}
\usepackage{courier}

\usepackage{color}
\usepackage{algorithm}
\usepackage{algorithmic}
\usepackage{amssymb}
\usepackage{amsmath}
\usepackage{graphicx,epsfig,subfigure,epstopdf}

\newtheorem{theorem}{Theorem}

\newtheorem{corollary}[theorem]{Corollary}
\newtheorem{assumption}[theorem]{Assumption}

\usepackage{algorithm, algorithmic}
\usepackage{graphicx} 
\usepackage{subfigure}
\graphicspath{{figs/}}

\usepackage{url}

\def\E{\mathbb{E}}
\def\R{\mathbb{R}}
\def\la{\langle}
\def\ra{\rangle}

\def\ifdtl{\ifthenelse{\equal{\dtl}{y}}}

\def\-{\!-\!}
\def\+{\!+\!}

\def \l {\lambda}

\def\eps{\epsilon}

\def\tw{\tilde{w}}
\def\bw{\bar{w}}

\def\md{\mathcal{D}}
\def\mb{\mathcal{B}}

\def\nF{\nabla F}
\def\nf{\nabla f}
\def\l({\left(}
\def\r){\right)}
\def\l[{\left[}
\def\r]{\right]}
\def\hw{\hat{w}}

\newcommand{\bigO}{\mathcal{O}}

\begin{document}
%
\title{Asynchronous Distributed Semi-Stochastic Gradient Optimization}
\author{
Ruiliang Zhang\quad
Shuai Zheng\quad
James T. Kwok\\
Department of Computer Science and Engineering\\
Hong Kong University of Science and Technology\\
Hong Kong\\
\{rzhangaf, szhengac, jamesk\}@cse.ust.hk
}
\maketitle
\begin{abstract}
With the recent proliferation of large-scale learning problems, there have been a lot of
interest on distributed machine learning 
algorithms, particularly those that are 
based  on stochastic gradient descent (SGD) and its variants.
However, existing 
algorithms either suffer from slow convergence 
due to the inherent variance of stochastic gradients,
or have a fast linear convergence rate 
but at the expense of poorer solution quality.
In this paper, we combine their merits  by
proposing a 
fast 
distributed asynchronous SGD-based algorithm 
with variance reduction.
A constant learning rate
can be used,
and it is also guaranteed
to converge linearly to the optimal solution. Experiments on the Google Cloud Computing
Platform demonstrate that the proposed algorithm
outperforms state-of-the-art 
distributed asynchronous algorithms
in terms of both wall clock time and solution quality.  
\end{abstract}

\section{Introduction}

With the recent proliferation of big data, 
learning the parameters in a large machine learning model is a
challenging problem.
A popular approach 
is to use
stochastic gradient descent (SGD) and
its variants
\cite{bottou2010large,dean2012large,gimpel2010distributed}. 
However, it can still be difficult to store and process 
a big data set
on one single machine. Thus,  there is now growing interest in
distributed machine learning algorithms.
\cite{dean2012large,gimpel2010distributed,niu2011hogwild,shamir2014communication,zhang2014asynchronous}.
The data set is partitioned into subsets, assigned to multiple machines, and the optimization
problem is solved
in a distributed manner.

In general, distributed architectures can be categorized as shared-memory \cite{niu2011hogwild} or
distributed-memory \cite{dean2012large,gimpel2010distributed,shamir2014communication,zhang2014asynchronous}.
In this paper, we will focus on the latter, which is more
scalable.
Usually, one of the machines is the {\em server}, while the rest are {\em workers}.
The workers store the data subsets, perform local computations and send their updates to the
server. The server then aggregates the local information, performs the actual update on the
model parameter, and sends it back to the workers. Note that workers only need to communicate with the server but not among them. 
Such a distributed computing model has been commonly used in many recent large-scale machine learning
implementations
\cite{dean2012large,gimpel2010distributed,shamir2014communication,zhang2014asynchronous}.

Often,
machines 
in these systems have to 
run synchronously 
\cite{boyd2011distributed,shamir2014communication}.
In each iteration, information from all workers 
need to be ready
before the server can 
aggregate the updates.
This can be expensive due to communication overhead and random network delay. It also suffers
from the straggler problem \cite{albrecht2006loose}, in which the system can move forward only at the pace of the slowest worker.

To alleviate these problems, asynchronicity is introduced 
\cite{agarwal2012distributed,dean2012large,ho2013more,li2014communication,zhang2014asynchronous}.
The server 
is allowed to 
use only staled (delayed) information from the workers,
and thus only needs to wait for a much smaller number of workers in each iteration.
Promising theoretical/empirical results have been reported.
One prominent example of asynchronous SGD is the {\em downpour SGD}
\cite{dean2012large}. 
Each worker 
independently
reads the parameter from the
server, computes the local gradient, and sends it back
to the server. The server then immediately updates the parameter using the worker's gradient
information.
Using an adaptive learning rate \cite{duchi2011adaptive}, 
downpour SGD
achieves state-of-the-art performance. 

However, in order for these algorithms to converge,
the learning rate has to decrease not only with the number of iterations
(as in standard single-machine SGD algorithms \cite{bottou2010large}),
but also with the maximum delay 
$\tau$
(i.e., the duration between the time the gradient is computed by the worker and it is used by
the server)
\cite{ho2013more}.
On the other hand, note that downpour SGD does not constrain 
$\tau$, but no convergence guarantee is provided.

In practice, a decreasing learning rate leads
to slower convergence \cite{bottou2010large,johnson2013accelerating}.
Recently, 
\citeauthor{feyzmahdavian2014delayed} 
(\citeyear{feyzmahdavian2014delayed})
proposed the \emph{delayed proximal
gradient} method in which the delayed gradient is used to update an analogously delayed model
parameter (but not its current one). It is shown that even with 
a constant learning rate, the algorithm converges linearly to within $\epsilon$ of the optimal
solution. However, to achieve a small $\epsilon$, the learning rate needs to be small, which
again means slow convergence.

Recently, there has been the flourish development of variance reduction techniques for SGD.
Examples include 
{\em stochastic average gradient} 
(SAG) \cite{roux2012stochastic}, 
{\em stochastic variance reduced gradient} 
(SVRG) \cite{johnson2013accelerating},
{\em minimization by incremental surrogate optimization} (MISO)
\cite{Mairal2013,mairal2014incremental}, 
SAGA \cite{defazio2014saga}, 
{\em stochastic dual coordinate descent}
(SDCA) \cite{shalev2013stochastic}, and 
Proximal SVRG \cite{xiao2014proximal}.
The idea is to use past gradients to progressively reduce the stochastic gradient's variance, so that a constant
learning rate can again be used. 
When the optimization objective is strongly convex and Lipschitz-smooth, all these
variance-reduced SGD algorithms converge linearly to the optimal solution.
However, their space requirements are different. In particular,
SVRG is advantageous in that it only needs to store the averaged sample
gradient, while SAGA and SAG have to store
all the samples' most recent gradients.
Recently, 
\citeauthor{mania2015perturbed}
(\citeyear{mania2015perturbed})
and
\citeauthor{reddi2015variance} 
(\citeyear{reddi2015variance})
extended
SVRG to the parallel asynchronous setting. Their algorithms are
designed for shared-memory multi-core systems, and assume that the
data samples are sparse. However, 
in a distributed computing environment,  the samples need to be
mini-batched to reduce the communication overhead between workers and server. 
Even when the samples are sparse,  the resultant mini-batch
typically is not.

In this paper, we propose a distributed asynchronous SGD-based  algorithm with variance
reduction, and
the data samples can be sparse or dense.
The algorithm is easy to implement, highly scalable,
uses a constant learning rate,
and converges linearly to the optimal solution.
A prototype is implemented on the Google Cloud Computing Platform.
Experiments on several big data sets from the Pascal Large Scale Learning
Challenge and LibSVM archive
demonstrate that it outperforms the state-of-the-art.

The rest of the paper is organized as follows. We first introduce related work.
Next, we present the proposed distributed asynchronous algorithm.
This is then
followed by 
experimental results including comparisons with the state-of-the-art 
distributed asynchronous algorithms, and the last section gives concluding
remarks.


\section{Related Work}

Consider the following optimization problem
\begin{equation} \label{F}
\min_w F(w) \equiv \frac{1}{N} \sum_{i=1}^N f_i(w).
\end{equation}
In many machine learning applications, $w\in \R^d$ is the model parameter,
$N$ is the number of training samples, and
each $f_i:\R^d\rightarrow \R$ is the
loss 
(possibly regularized) 
due to sample $i$.
The following assumptions are commonly made.
\begin{assumption}
\label{smooth}
Each $f_i$ is $L_i$-smooth \cite{nesterov2004introductory}, i.e., 
$f_i(x) \le f_i(y) + \la \nf_i(y),x-y\ra +\frac{L_i}{2} \|x-y\|^2 \;
\forall x, y$.
\end{assumption}
\begin{assumption} \label{strongly_convex}
$F$ is $\mu$-strongly convex \cite{nesterov2004introductory}, i.e., 
$F(x) \ge F(y) + \la F(y),x-y\ra +\frac{\mu}{2} \|x-y\|^2 \;
\forall x, y$.
 \end{assumption}


\subsection{Delayed Proximal Gradient (DPG)}
\label{sec:dpg}

At iteration $t$ of the DPG
\cite{feyzmahdavian2014delayed}, a worker uses $w^{t-\tau_t}$, the copy  of $w$
delayed by $\tau_t$ iterations,
to compute the stochastic gradient 
$g^{t-\tau_t}=\nabla f_i(w^{t-\tau_t})$
on a random sample $i$.
The delayed gradient is used to update the correspondingly delayed parameter copy
$w^{t-\tau_t}$ to
$\hw^{t-\tau_t}=w^{t-\tau_t}-\eta g^{t-\tau_t}$,
where $\eta$ is a constant learning rate.
This
 $\hw^{t-\tau_t}$ is then 
sent to the server,
which obtains the new iterate $w^{t+1}$ as 
a convex combination of the current $w^t$ and
$\hw^{t-\tau_t}$:
\begin{equation} \label{eq:delayed_proximal_grad}
 w^{t+1}=(1-\theta)w^t+\theta\hw^{t-\tau_t},
\;\; \theta\in(0,1]. 
\end{equation}

It can be shown that
the $\{w^t\}$ sequence converges linearly to the optimal solution $w^*$, but only  within a tolerance of
$\eps$, i.e.,
\[ \E \left[ F(w^t)-F(w^*) \right]\le \rho^t(F(w^0)-F(w^*)) +\epsilon, \]
for some $\rho<1$ and
$\epsilon>0$. 
The tolerance $\epsilon$ can be reduced by reducing $\eta$, though at the expense
of increasing $\rho$ and thus slowing down convergence. 
Moreover, though the learning rate of DPG is typically larger than that of SGD, the gradient
of DPG (i.e., $\hw^{t-\tau_t}$)
is delayed and slows convergence.


\subsection{Stochastic Variance Reduced Gradient}
 
The SGD, 
though simple and scalable,
has a 
slower  
convergence rate
than
batch gradient descent \cite{Mairal2013}.
As noted in \cite{johnson2013accelerating}, the underlying reason 
is that the stepsize of SGD has to be
decreasing so as to control the gradient's variance.
Recently, by observing 
that the training set is always finite in practice, a number of techniques have been developed
to reduce this variance and thus allows the use of a constant stepsize
\cite{defazio2014saga,johnson2013accelerating,Mairal2013,roux2012stochastic,xiao2014proximal}.

In this paper, we focus on one of  most popular techniques in this family, namely
the 
{\em stochastic variance reduction gradient} (SVRG)
\cite{johnson2013accelerating}
(Algorithm \ref{alg:svrg}).
It is advantageous 
in that 
no extra space 
is needed
for the intermediate gradients or dual variables.
The algorithm proceeds in stages.
At the beginning of each stage, 
the gradient
$\nF(\tw) = \frac{1}{N}\sum_{i=1}^N\nabla f_i(\tw)$
is computed on the whole data set 
using a past parameter estimate $\tw$ (which is updated across stages).
For each 
subsequent
iteration $t$ in this stage,
the approximate gradient 
\[ \hat{\nabla} f_i(w^t) =\nabla f_i(w^t) - \nabla f_{i}(\tw) + \nF(\tw) \]
is used,
where $i$ is a sample randomly selected from $\{1, 2, \dots, N\}$.
Even with a constant learning rate $\eta$, 
the (expected) variance of $\hat{\nabla} f(w^t)$
goes to zero progressively, and
the algorithm achieves linear convergence.

\begin{algorithm}[htp]
\caption{Stochastic variance reduced gradient (SVRG) \cite{johnson2013accelerating}.}
\label{alg:svrg}
\begin{algorithmic}[1]
\STATE {\bf Initialize} $\tw^0$;
\FOR{$s = 1,2,...$}
\STATE $\tw = \tw^{s-1}$;
\STATE $\nF(\tw) = \frac{1}{N}\sum_{i=1}^N \nf_i(\tw)$;
\STATE $w^0 = \tw$;
\FOR{$t=0,1,\dots,m-1$}
\STATE randomly pick $i\in\{1,\dots,N\}$;
\STATE $w^{t+1}=w^t-\eta
\hat{\nabla} f_i(w^t)$;
\ENDFOR
\STATE set $\tw^s=w^t$ for a randomly chosen $t\in\{0,\dots,m-1\}$;
\ENDFOR
\end{algorithmic}
\end{algorithm}

In contrast to DPG, SVRG can converge to the optimal solution.
However, 
though SVRG has been extended to the parallel asynchronous setting on shared-memory multi-core
systems \cite{reddi2015variance,mania2015perturbed},
its use and convergence properties 
in a distributed asynchronous learning setting
remain unexplored.


\section{Proposed Algorithm}

In this section, we consider
the distributed asynchronous setting, and
propose a hybrid of an improved DPG algorithm and SVRG that inherits the advantages of both.
Similar to the two algorithms, it also uses a constant learning rate
(which is typically larger than the one used by SGD), but with guaranteed linear convergence to the optimal solution.


\subsection{Update using Delayed Gradient}

We replace the 
SVRG update (line~8
in Algorithm~\ref{alg:svrg}) by
\begin{equation} \label{ours}
w^{t+1} = (1-\theta)v^t +\theta \bw^{t-\tau_t},
\end{equation}
where 
\[ v^t=w^t-\eta\hat{\nabla} f_i(w^{t-\tau_t})\]
and
\begin{equation} \label{eq:bw}
\bw^{t-\tau_t}=w^{t-\tau_t}-\eta\hat{\nabla} f_i(w^{t-\tau_t}). 
\end{equation} 
Obviously, when $\tau_t=0$, (\ref{ours}) reduces to  standard SVRG. 
Note that 
both the parameter and gradient 
in $\bw^{t-\tau_t}$ 
are for the same iteration ($t-\tau_t$),
while 
$v^t$ is noisy as
the gradient is delayed
(by $\tau_t$).
This delayed gradient cannot be too old. Thus, 
similar to \cite{ho2013more},
we impose the \emph{bounded delay} condition that $\tau_t \leq \tau$ for some $\tau>0$.
This $\tau$ parameter determines the maximum duration between the time the gradient is computed and till it is used.
A larger
$\tau$ allows  more asynchronicity,  but also adds noise to the gradient and
thus may slow convergence.

Update (\ref{ours}) is similar to (\ref{eq:delayed_proximal_grad}) in DPG, but with  
two
important differences. First,
the gradient $\nabla f_i(w^{t-\tau_t})$  in DPG
is replaced
by its variance-reduced counterpart $\hat{\nabla} f_i(w^{t-\tau_t})$.
As will be seen, this allows convergence to the optimal solution using a constant learning rate.
The second difference is that 
the delayed gradient 
$\hat{\nabla} f_i(w^{t-\tau_t})$ 
is used not only on 
the past iterate $w^{t-\tau_t}$, but also on 
the current iterate $w^t$.
This can potentially yield faster progress, as is most apparent 
when
$\theta=0$. 
In this special case,
DPG reduces to $w^{t+1} =w^t$, and makes no progress; while (\ref{ours}) reduces to
the asynchronous SVRG update in 
\cite{reddi2015variance}.


\subsection{Mini-Batch}

In a distributed algorithm, communication overhead is incurred when a worker pulls
parameters from 
the server
or pushes update to it. In a distributed SGD-based algorithm, the communication cost is
proportional to the number of gradient evaluations made by the workers.
Similar
to the other SGD-based
distributed algorithms \cite{gimpel2010distributed,dean2012large,ho2013more}, this cost can be
reduced by the use of a mini-batch.  Instead of pulling parameters from the server after every sample, the
worker pulls only after processing each 
mini-batch of size $B$.


\subsection{Distributed Implementation}

There is a {\em scheduler}, a {\em server} and  $P$ {\em workers}.
The server keeps a clock (denoted by an integer $t$), the most updated copy of
parameter $w$,
a past parameter estimate $\tw$ and the corresponding full gradient 
$\nF(\tw)$ evaluated on the whole training set
$\md$ (with $N$ samples). 
We divide $\md$ into $P$ disjoint subsets $\md_1,\md_2,\dots,\md_P$,
where $\md_p$ is
owned by worker $p$.
The number of samples in $\md_p$ is denoted $n_p$.
Each worker $p$ also keeps a local copy 
$\tw_p$
of $\tw$.

In the following,
a \emph{task} refers to an event timestamped by 
the scheduler. 
It can be issued by the scheduler
or a worker, and received by either the server or workers. 
Each worker can only process one task at a time.
There are two types of tasks, \emph{update task} and \emph{evaluation task}, and will be
discussed in more detail in the sequel. A worker may 
pull the parameter
from the server by
sending a \emph{request}, which carries the type and timestamp of the task being run by the
worker.


\subsubsection{Scheduler}

The scheduler
(Algorithm~\ref{alg:scheduler})
runs in stages.
In each stage, it first issues
$m$ update tasks to the workers, where $m$ is usually a multiple of $\lceil N/B\rceil$ 
as in SVRG.
After spawning enough tasks, the server
measures the progress
 by issuing an evaluation task to the server and all workers.
As will be seen, the server ensures that evaluation is carried out only after all update tasks for the current stage have
finished.
If 
the stopping condition
is met,
the scheduler informs the server and all workers
by issuing a STOP
command; otherwise, it moves to the next stage and sends more update tasks.

\begin{algorithm}[htp]
\caption{Scheduler.}
\label{alg:scheduler}
\begin{algorithmic}[1]
\FOR{$s = 1,\dots,S$}
\FOR{$k=1,\dots,m$}
\STATE pick worker $p$ with probability $\frac{n_p}{N}$;
\STATE issue an update task to the worker with timestamp $t=(s-1)m+k$;
\ENDFOR
\STATE issue an evaluation task (with timestamp $t=sm+1$) to workers and server;
\STATE wait and collect progress information from workers;
\IF {progress meets stopping condition}
\STATE issue a STOP command to the workers and server;
\ENDIF
\ENDFOR
\end{algorithmic}
\end{algorithm}


\subsubsection{Worker}
\label{sec:worker}

At stage $s$, when worker $p$ receives an \emph{update} task with timestamp $t$, it sends a parameter pull
request to the server.  This request will not be responded by the server until it finishes all
tasks with timestamps before $t-\tau$.

Let $\hw_{p,t}$ be the parameter value pulled.
Worker $p$ selects a mini-batch $\mb^t\subset \md_p$ (of size $B$) randomly from its local
data set. 
Analogous to  $\bw^{t-\tau_t}$ in (\ref{eq:bw}), it computes
\begin{equation} \label{intermediate_update}
\bw_{p,t} = \hw_{p,t}-\eta  \Delta w_{p,t},
\end{equation}
where $\Delta w_{p,t}$ is the mini-batch gradient evaluated at
$\hw_{p,t}$.
An update task is then issued
to push
$\bw_{p,t}$ and $\Delta w_{p,t}$ to the server.

When a worker receives an \emph{evaluation} task, 
it again sends a parameter pull request to the server.
As will be seen
in the following section,
the pulled $\hw_{p,t}$ 
will always be the latest $w$ kept by the server in the current stage.
Hence, the $\hw_{p,t}$'s pulled by all workers are the same. 
Worker $p$ then updates
$\tw_p$ as $\tw_p=\hw_{p,t}$,
computes and pushes the corresponding gradient
\[ \nF_p(\tw_p)=\frac{1}{n_p}\sum_{i\in\md_p}\nf_i(\tw_p) \]
to the server.
To inform the scheduler of its progress,
worker $p$ also computes its contribution  
to the optimization objective
$\sum_{i\in\md_p}f_i(\tw_p)$
and
pushes it to the scheduler.
The whole worker procedure
is shown in Algorithm~\ref{alg:worker}.

\begin{algorithm}[htp]
\caption{Worker $p$ receiving an update/evaluation task $t$ at stage $s$.}
\label{alg:worker}
\begin{algorithmic}[1]
\STATE send a parameter pull request to the server;
\STATE wait for response from the server;
\IF {task $t$ is an update task}
\STATE pick a mini-batch subset $\mb^t$ randomly from the local data set;
\STATE compute mini-batch gradient $\Delta w_{p,t}$ 
and $\bw_{p,t}$ using 
(\ref{intermediate_update}), and
push them to the server as an update task;
\ELSE
\STATE set $\tw_p=\hw_{p,t}$;
\COMMENT{task $t$ is an evaluation task}
\STATE push the local subset gradient $\nF_p(\tw_p)$
to the server as an update task;
\STATE push the local objective value to the scheduler;
\ENDIF
\end{algorithmic}
\end{algorithm}


\subsubsection{Server}
\label{server}

There are two threads running on the server. One is a daemon thread that 
responds to parameter pull requests from workers
(Algorithm~\ref{alg:daemon});
and the other is a computing thread for handling
update tasks from workers and evaluation tasks from the scheduler
(Algorithm~\ref{alg:compute}).

When the daemon
thread receives a parameter pull request, it reads the
type and
timestamp $t$ within.
If the request is 
from a worker running an update task,
it checks whether all update tasks before $t-\tau$
have finished. If not, the request remains in the buffer;
otherwise, it
pushes its $w$ value to the requesting worker. Thus, this controls the allowed asynchronicity. On the other hand,
if the request is 
from a worker executing an evaluation task,
the
daemon thread does not push $w$ to the workers until all update tasks
before $t$ have finished.
This ensures that the $w$ pulled by the worker is the most up-to-date for the current stage.

\begin{algorithm}[htp]
\caption{Daemon thread of the server.}
\label{alg:daemon}
\begin{algorithmic}[1]
\REPEAT
\IF {pull request buffer is not empty}
\FOR {{\bf each} request with timestamp $t$ in the buffer}
\IF {request is triggered by an update task}
\IF {all update tasks before $t-\tau$ have finished}
\STATE push $w$ to the requesting worker;
\STATE remove request from buffer; 
\ENDIF
\ELSE
\IF {all update tasks before $t$ have finished}
\STATE push $w$ to the requesting worker;
\STATE remove request from buffer; 
\ENDIF
\ENDIF
\ENDFOR
\ELSE
\STATE sleep for a while;
\ENDIF
\UNTIL STOP command is received.
\end{algorithmic}
\end{algorithm}

When the computing thread receives an update task (with timestamp $t$) from worker $p$, 
the $\bw_{p,t}$ and $\Delta w_{p,t}$ contained inside are read.
Analogous to (\ref{ours}),
the server updates  $w$
as
\begin{equation} \label{full_upate}
\begin{aligned}
w & = (1-\theta)(w - \eta \Delta w_{p,t}) + \theta \bw_{p,t},
\end{aligned}
\end{equation}
and marks this task as finished. During the update,
the computing thread locks $w$
so that the daemon thread cannot access until the update
is finished. 

When the server receives an evaluation task, it synchronizes all workers,  and sets
$\tw = w$.
As all $\tw_p$'s are the same and equal to $\tw$, one can 
simply aggregate the local gradients 
to obtain 
$\nF(\tw) = \sum_{p=1}^P q_p \nF_p(\tw_p)$,
where $q_p=\frac{n_p}{N}$.
The server then broadcasts $\nF(\tw)$ to all workers.

\begin{algorithm}[t]
\caption{Computing thread of the server.}
\label{alg:compute}
\begin{algorithmic}[1]
\REPEAT
\STATE wait for tasks;
\IF {an update task received}
\STATE update $w$ using (\ref{full_upate}),
and mark this task as finished;
\ELSE
\STATE wait for all update tasks to finish;
\STATE set $\tw=w$;
\STATE collect local full gradients from workers and update $\nF(\tw)$;
\STATE broadcast $\nF(\tw)$ to all workers;
\ENDIF
\UNTIL 
STOP command is
received.
\end{algorithmic}
\end{algorithm}

\subsection{Discussion}

Two state-of-the-art distributed asynchronous SGD algorithms are the downpour SGD
\cite{dean2012large} and Petuum SGD \cite{ho2013more,dai2013petuum}.  Downpour SGD does not impose
the bounded delay condition (essentially, $\tau=\infty$), while Petuum SGD does.
Note that there is a subtle difference in the bounded delay condition of the proposed algorithm
and that of Petuum SGD.
In Petuum SGD, the amount of staleness is measured between workers, namely that
the slowest and fastest workers must be less than $s$ timesteps apart (where $s$ is the staleness
parameter).
Consequently, the delay 
in the gradient 
is always a multiple of $P$ and is upper-bounded by $sP$.
On the other hand, in the proposed algorithm, the bounded delay condition is imposed on the update tasks.
It can be easily seen that $\tau$ is also the maximum delay in the gradient. Thus,
$\tau$ can be any number which is not necessarily a multiple of $P$.


\subsection{Convergence Analysis}
\label{sec:conv}

For simplicity of analysis, we assume that the mini-batch size is one.
Let $\tw^S$ be the $w$ stored in the server at the end of stage $S$
(step~7 in
Algorithm~\ref{alg:compute}).
The following Theorem shows 
linear convergence
of the proposed algorithm. 
It is the first  such result for distributed asynchronous SGD-based algorithms with constant learning rate.

\begin{theorem} \label{theorem1}
Suppose that problem (\ref{F}) satisfies Assumption~\ref{smooth} and \ref{strongly_convex}.
Let $L=\max\{L_i\}_{i=1}^N$, and
$\gamma=
\left(1-2\eta(\mu-\frac{\eta L^2}{\theta})\right)^{\frac{m}{1+\tau}}+ \frac{\eta
L^2}{\theta\mu-\eta L^2}$.
With $\eta\in(0,\frac{\mu\theta}{2L^2})$ 
and $m$ sufficiently large such that $\gamma<1$. Assume that the scheduler has been run for $S$ stages, 
we obtain the following linear rate:
\[ \E[F(\tw^S)-F(w^*)]  \le \;\;\gamma^S[F(\tw^0)-F(w^*)]. \]
\end{theorem}

As $L > \mu$, it is easy to see that  $1-2\eta(\mu-\frac{\eta L^2}{\theta})$ and $\frac{\eta L^2}{\theta\mu-\eta L^2}$ are both smaller than 1. Thus, $\gamma<1$ can be guaranteed for a sufficiently large $m$.
Moreover, as $F$ is strongly convex,  
the following 
Corollary  
shows  that
$\tw^S$ also converges to 
$w^*$.
In contrast, DPG only converges to within a tolerance of $\eps$.

\begin{corollary}
$\E \|\tw^S-w^*\|^2\leq 2\gamma^S
[F(\tw^0)-F(w^*)]/ \mu$.
\end{corollary}

When $\tau < P$, the server can serve at most $\tau$ workers simultaneously. For maximum parallelism,
$\tau$ should increase with $P$.
However, 
$\gamma$ 
also increases
with $\tau$.
Thus,
a larger $m$ 
and/or $S$  may be needed to achieve the same solution quality.

Similar to DPG,
our learning rate does not depend on the
delay $\tau$ and the number of workers $P$. This learning rate can be significantly larger than the
one in Pentuum SGD \cite{ho2013more}, which has to be decayed and is proportional
to $\bigO(1/\sqrt{P})$. Thus, 
the proposed algorithm can be much faster,
as will be confirmed in the experiments. 
While our bound may be loose due to the use of worst-case analysis, linear convergence is always
guaranteed for any $\theta \in(0,1]$.


\section{Experiments}

In this section, we consider 
the $K$-class logistic regression problem:
\[\min_{\{w_k\}_{k=1}^K} \frac{1}{N} \sum_{i=1}^N\sum_{k=1}^K - I(y_i=k)\log\left(\frac{\exp(
w_k^T
x_i
)}{\sum_{j=1}^K \exp(
w_j^T
x_i
)}\right),\] 
where 
$\{(x_i, y_i)\}_{i=1}^N$ are the training samples, $w_k$ is the parameter vector of class $k$, and $I(\cdot)$ is the indicator
function which returns 1 when the argument holds, and 0 otherwise.
Experiments are performed on the 
\emph{Mnist8m} 
and 
\emph{DNA} 
data sets (Table~\ref{table:data set})
from the 
\emph{LibSVM} archive\footnote{\url{https://www.csie.ntu.edu.tw/~cjlin/libsvmtools/data sets/}}
and
Pascal Large Scale Learning Challenge\footnote{\url{http://argescale.ml.tu-berlin.de/}}.

\begin{table}[htbp]
\begin{center}
\begin{tabular}{c|c|c|c} \hline
& \#samples  &  \#features  & \#classes \\ \hline
{\em Mnist8m} & 8,100,000 & 784 & 10 \\ \hline
{\em DNA} & 50,000,000 & 800 & 2 \\ \hline
\end{tabular}
\vspace{-.1in}
\caption{Summary of the data sets used.}
\label{table:data set}
\end{center}
\end{table}

Using the Google Cloud Computing Platform\footnote{\url{http://cloud.google.com}},
we set up a cluster with 18 computing nodes.
Each node is a google cloud n1-highmem-8 instance
with eight cores and 52GB memory. Each scheduler/server takes one instance, while each worker takes
a core. Thus, we have a maximum of 128 workers.
The system is implemented in C++, with the ZeroMQ
package for
communication.


\subsection{Comparison with the State-of-the-Art}
\label{sec:cmpr}

In this section, the following distributed asynchronous algorithms are compared:
\begin{itemize}
\item Downpour SGD 
(``\emph{downpour-sgd}") 
\cite{dean2012large},
with the adaptive learning rate in Adagrad \cite{duchi2011adaptive};
\item Petuum SGD \cite{dai2013petuum} (``\emph{petuum-sgd}"), the state-of-the-art
implementation of asynchronous SGD.
The learning rate is reduced by a fixed factor 0.95 at the end of each epoch.
The staleness $s$
is set to $2$, and so the delay in the gradient is bounded by 2$P$.
\item DPG \cite{feyzmahdavian2014delayed} (``\emph{dpg}");
\item A variant of DPG (``\emph{vr-dpg}``),
in which the gradient in update (\ref{eq:delayed_proximal_grad}) is replaced by
its variance-reduced version;
\item The proposed 
``distributed variance-reduced stochastic
gradient decent'' (\emph{distr-vr-sgd})
algorithm.
\item A special case of 
\emph{distr-vr-sgd},
with $\theta=0$ (denoted
``\emph{distr-svrg}``).
This reduces to the asynchronous SVRG algorithm in \cite{reddi2015variance}.
\end{itemize}

We use 128 workers. To maximize parallelism, we fix $\tau$ to 128.
The Petuum SGD code is
downloaded
from \url{http://petuum.github.io/}, while
the other asynchronous algorithms 
are implemented
in C++ by reusing most of our system's codes.
Preliminary studies show that synchronous SVRG is much slower and so is not included for comparison.
For 
\emph{distr-vr-sgd} and
\emph{distr-svrg},
the number of stages is $S= 50$, and the number of iterations in each stage is $m=\lceil N/B\rceil$, where $B$ is about $10\%$ of  
each worker's local data set
size.
For fair comparison, the other algorithms are run for $mS$ iterations.
All other parameters are tuned by a validation set, which is $1\%$ of the data set.

\begin{figure}[!ht]
\centering
\subfigure[\emph{Mnist8m}.]{\includegraphics[width=4.1cm]{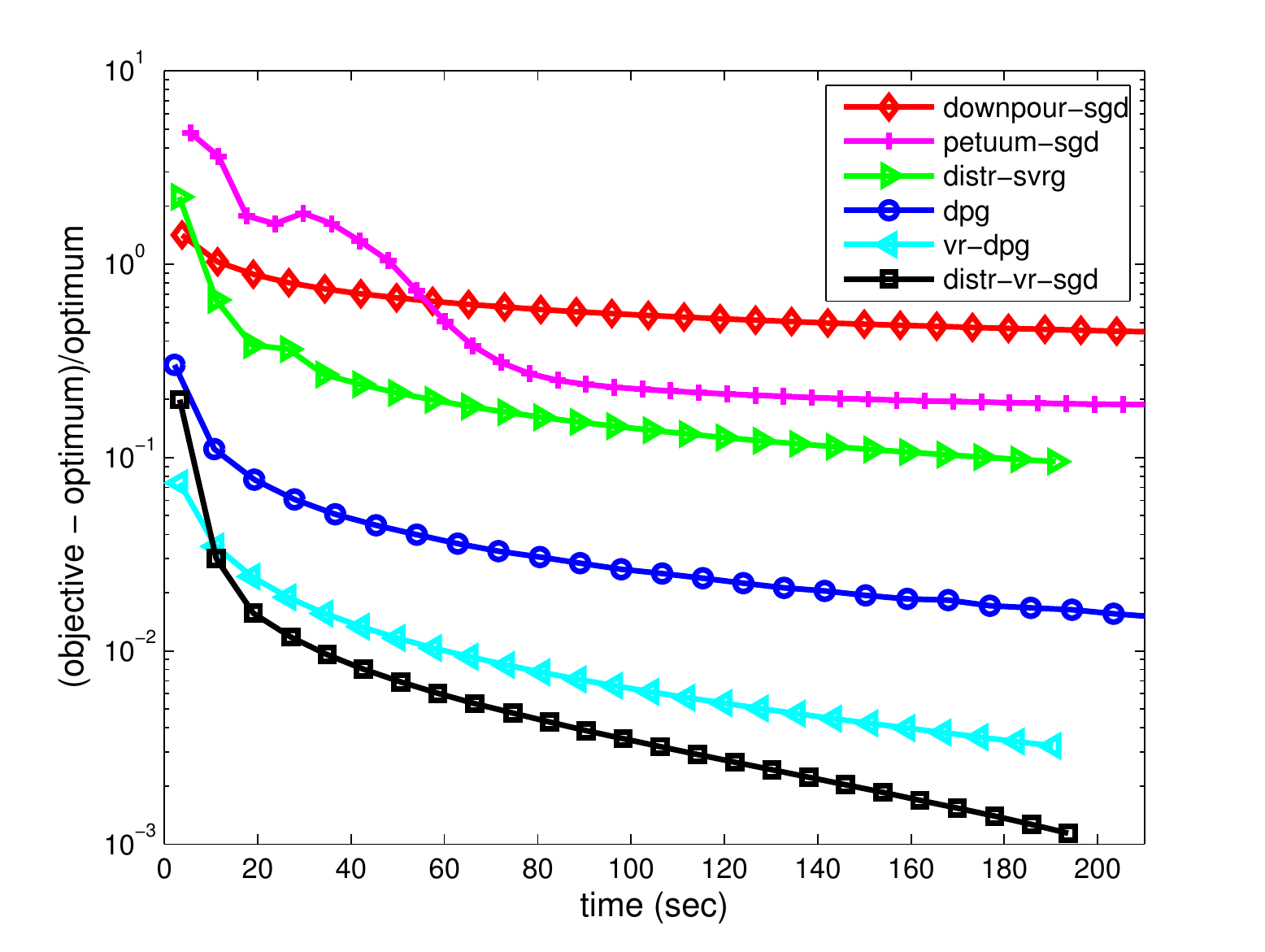}}
\subfigure[{\emph{DNA}}.]{\includegraphics[width=4.1cm]{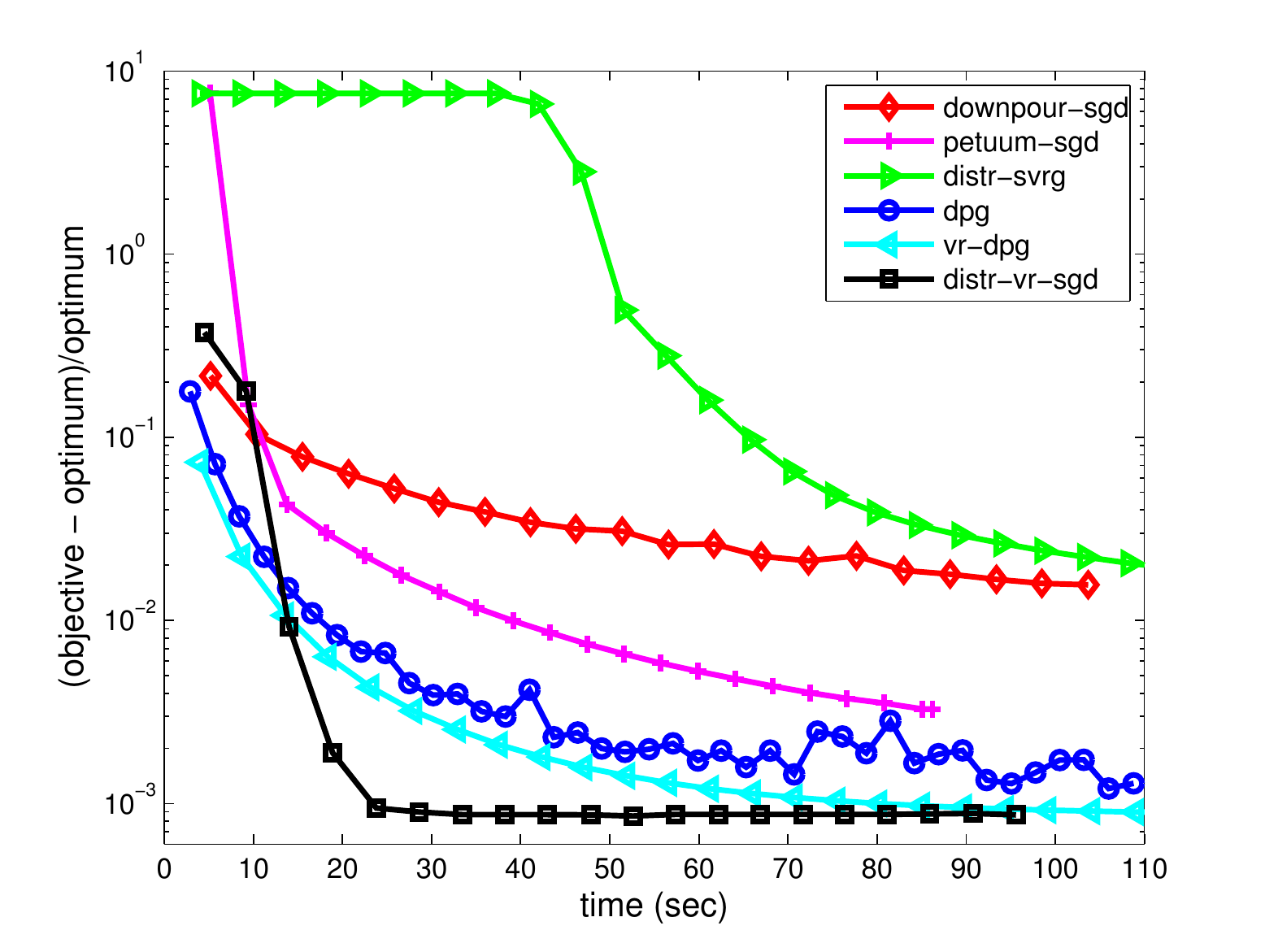}}
\vspace{-.1in}
\caption{Objective vs time (in sec).}
\label{fig:vs_sgd}
\end{figure}

Figure~\ref{fig:vs_sgd} shows convergence of the objective
w.r.t.
wall clock time. As can be seen,
\emph{distr-vr-sgd}
outperforms all the other algorithms. 
Moreover, 
unlike \emph{dpg},
it can converge to the optimal solution and attains a much smaller objective value.
Note that \emph{distr-svrg} is slow.
Since $\tau=128$,
the delayed gradient can be noisy, and the learning rate used by \emph{distr-svrg} (as determined by
the validation set) is small ($10^{-6}$ vs 
$10^{-3}$ in 
\emph{distr-vr-sgd}).
On the \emph{DNA} data set, \emph{distr-svrg} is even slower than
\emph{petuum-sgd} and \emph{downpour-sgd} (which use
adaptive/decaying learning rates). 
The \emph{vr-dpg},
which uses variance-reduced gradient, 
is always
faster than \emph{dpg}. 
Moreover,  
\emph{distr-vr-sgd} is faster than \emph{vr-sgd}, showing that replacing $w^t$ in
(\ref{eq:delayed_proximal_grad}) by $v^t$ in (\ref{ours}) is useful.

\subsection{Varying the Number of Workers}
\label{subsec:speedup}

\begin{figure}[ht]
\centering
\subfigure[\label{expt:scalability_obj}{\emph{Mnist8m}}.]{\includegraphics[width=4.1cm]
{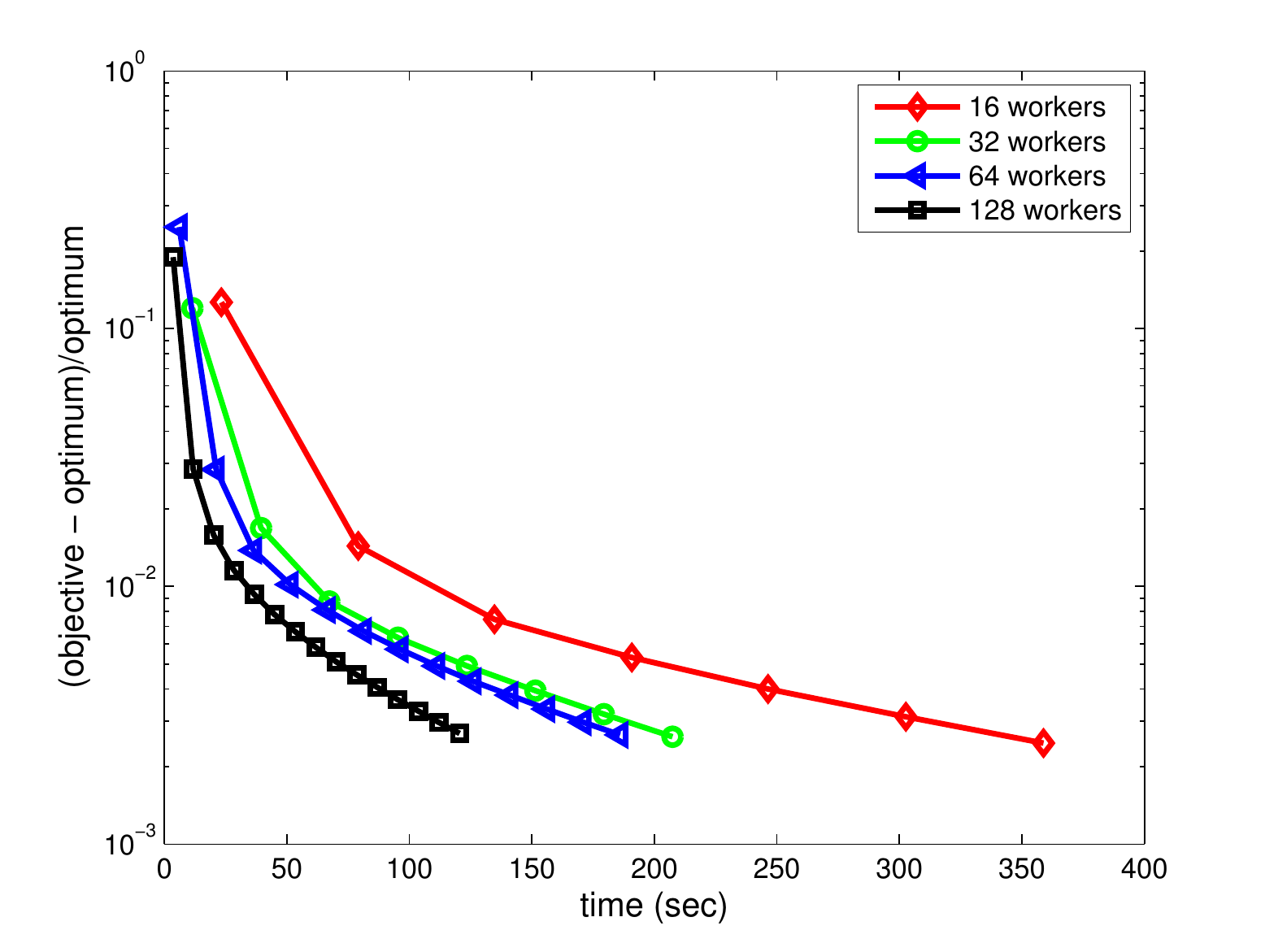}}
\subfigure[\label{expt:scalability_breakdown}{\emph{DNA}}.]{\includegraphics[width=4.1cm]{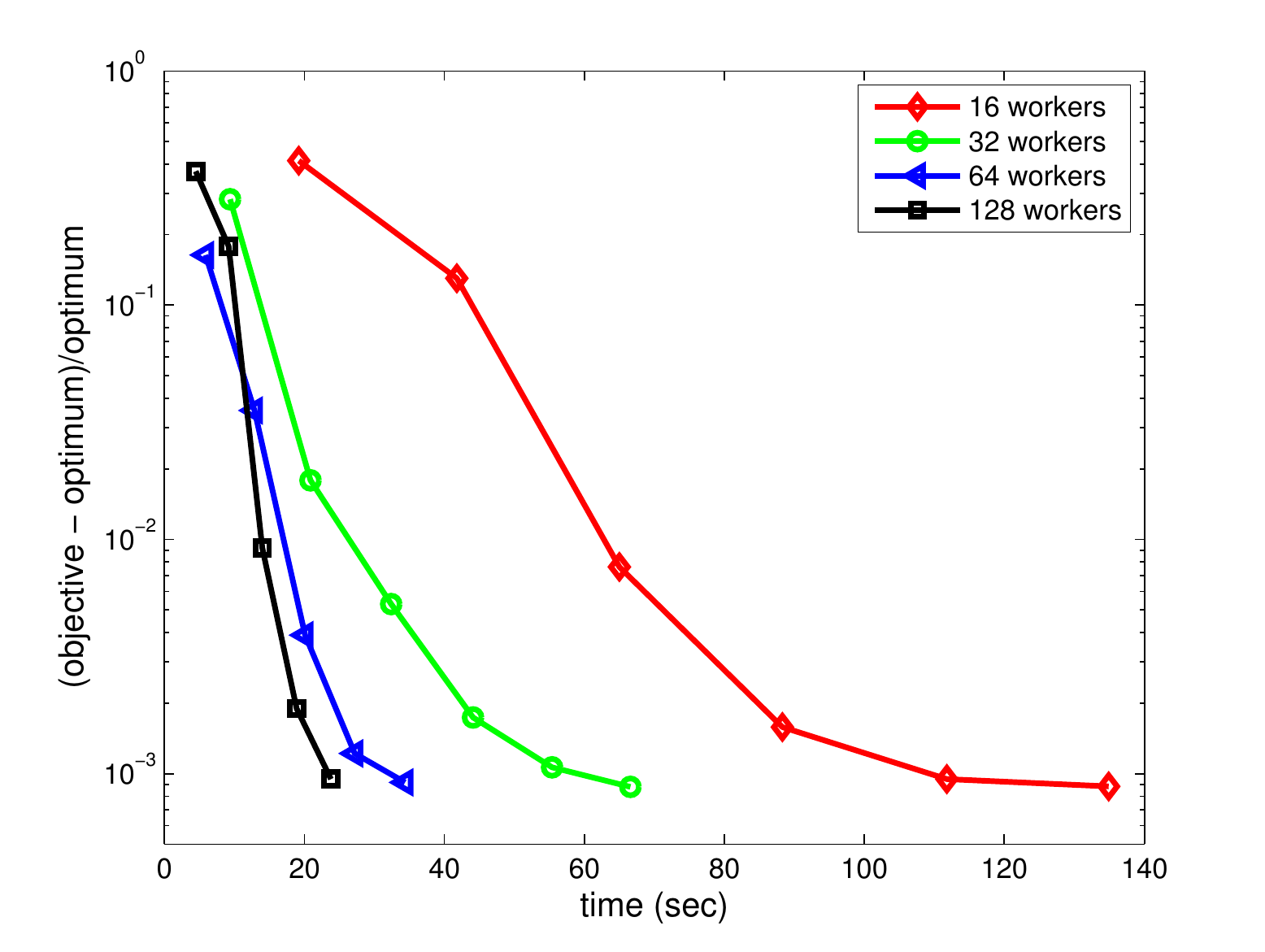}}
\vskip -.1in
\caption{Objective vs time (in sec), with different numbers of workers.}
\label{fig:scalability2}
\end{figure}

In this experiment, 
we 
run \emph{distr-vr-sgd} 
with varying number of workers (16, 32, 64 and 128)
until a target objective value is met.
Figure~\ref{fig:scalability2} shows convergence of the objective with time.
On the {\em Mnist8m} data set, using 128 workers is about 3 times faster than using 16 workers.  On the {\em DNA} data set, the speedup is about 6 times.

\begin{figure}[ht]
\centering
\subfigure[\label{expt:fixed_mnist}{\emph{Mnist8m}}.]{\includegraphics[width=4.1cm] {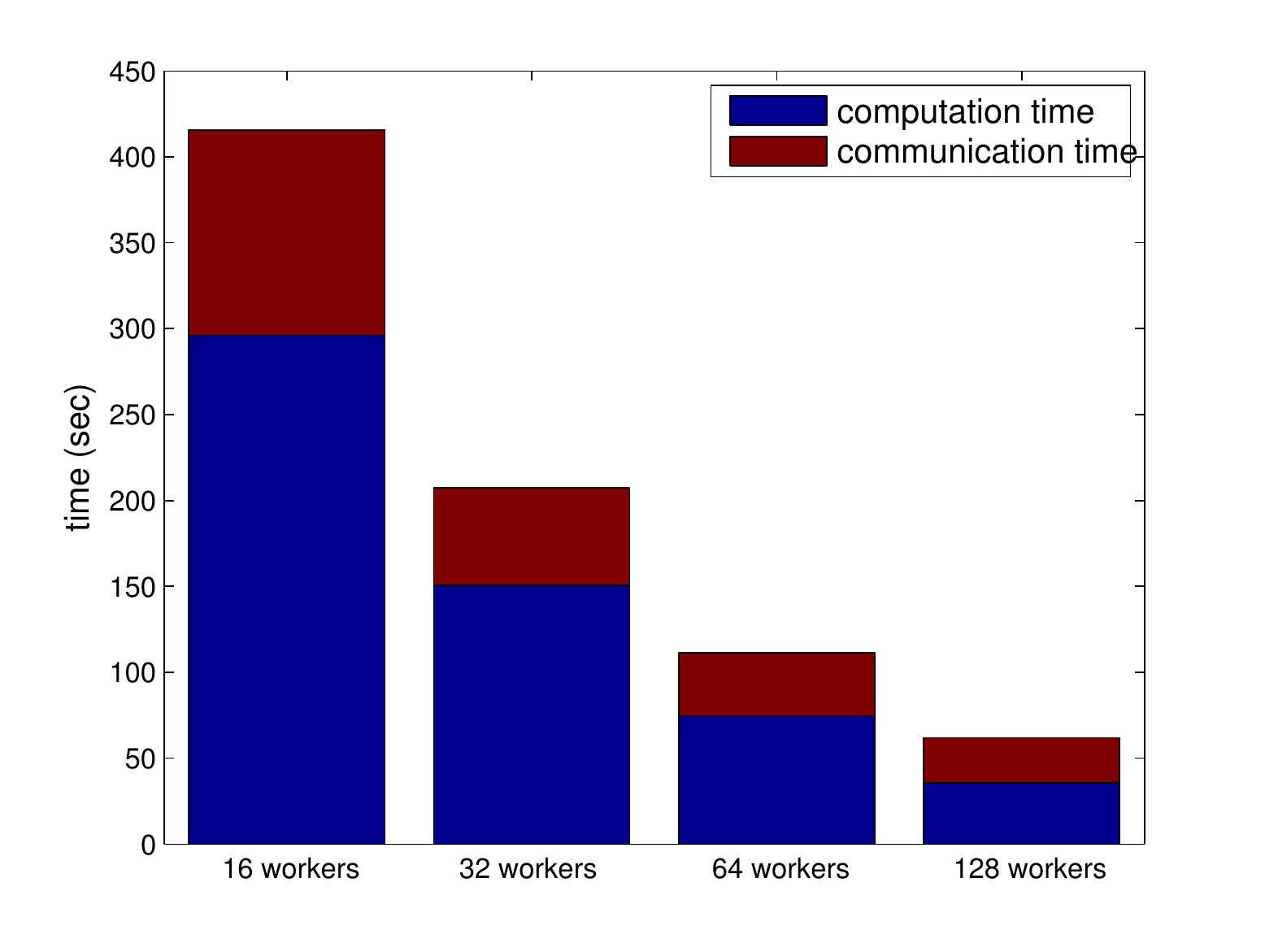}}
\subfigure[\label{expt:fixed_dna}{\emph{DNA}}.]{\includegraphics[width=4.1cm] {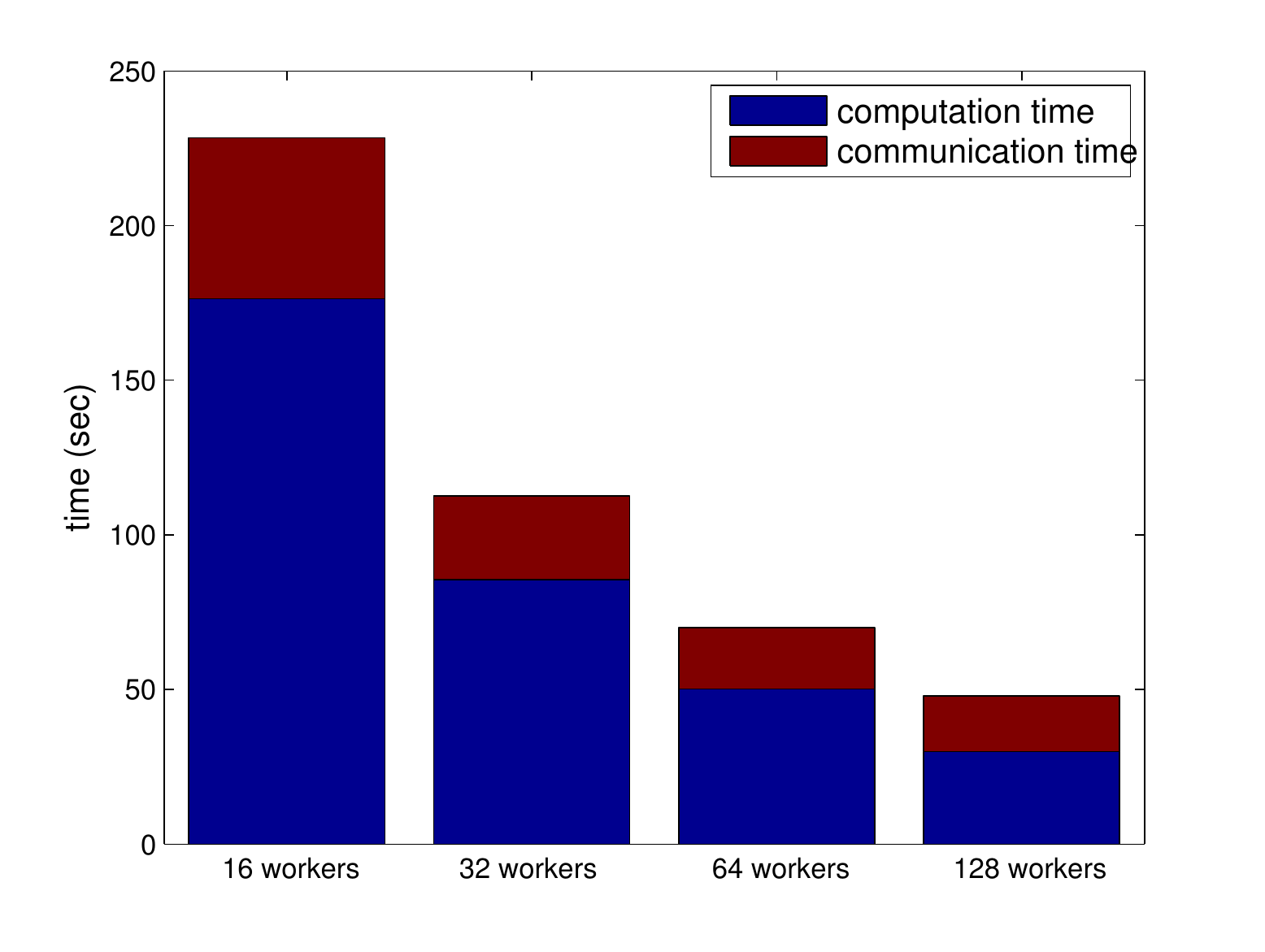}}
\vskip -.1in
\caption{Breakdown into computation time and communication time, with different numbers of workers.}
\label{fig:scalability1}
\end{figure}

Note that the most expensive step 
in the algorithm
is on gradient evaluations
(the scheduler and server operations are simple).
Recall that each stage has $m$ iterations, 
and each iteration involvs $O(B)$ gradient evaluations.
At the end of each stage, an additional $O(N)$ gradients are evaluated
to obtain the full gradient and monitor the progress. 
Hence, each worker spends $\bigO((mB+N)/P)$ time on computation. The computation
time thus decreases linearly with $P$, as can be seen from
the breakdown of total wall clock time into computation time and communication time
in Figure~\ref{fig:scalability1}.

Moreover, having more workers means
more tasks/data can be sent 
among the server and workers 
simultaneously,
reducing the communication time.
On the other hand, as
synchronization is required among all workers at the end of each stage, having more workers
increases the communication overhead. Hence, as can be seen from 
Figure~\ref{fig:scalability1},
the communication time first
decreases with the number of workers, but then increases as the communication cost in the synchronization step 
starts to  
dominate.


\subsection{Effect of $\tau$}

\begin{figure}[ht]
\centering
\subfigure[{\emph{Mnist8m}}.]{\includegraphics[width=4.1cm] {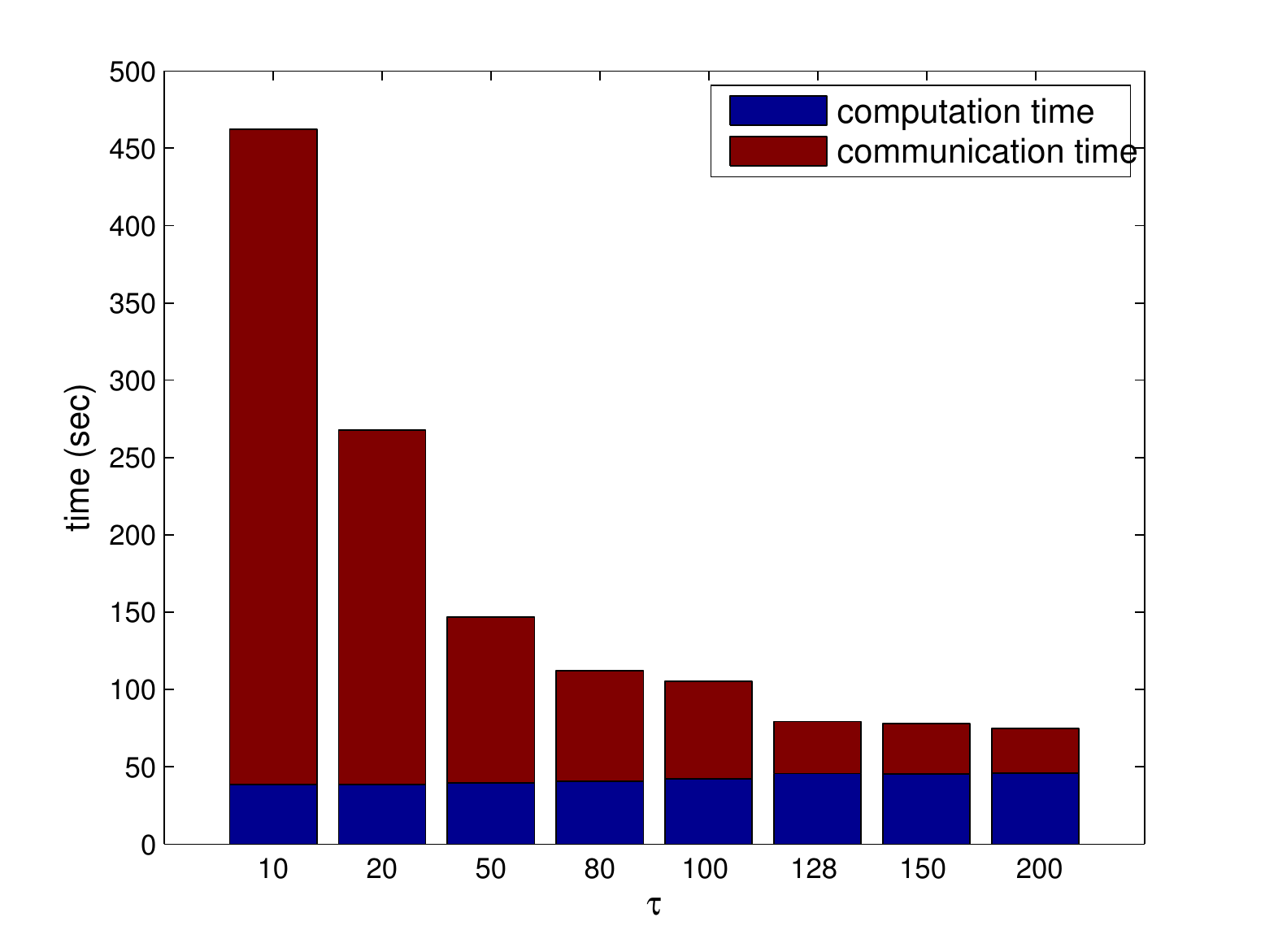}}
\subfigure[{\emph{DNA}}.]{\includegraphics[width=4.1cm] {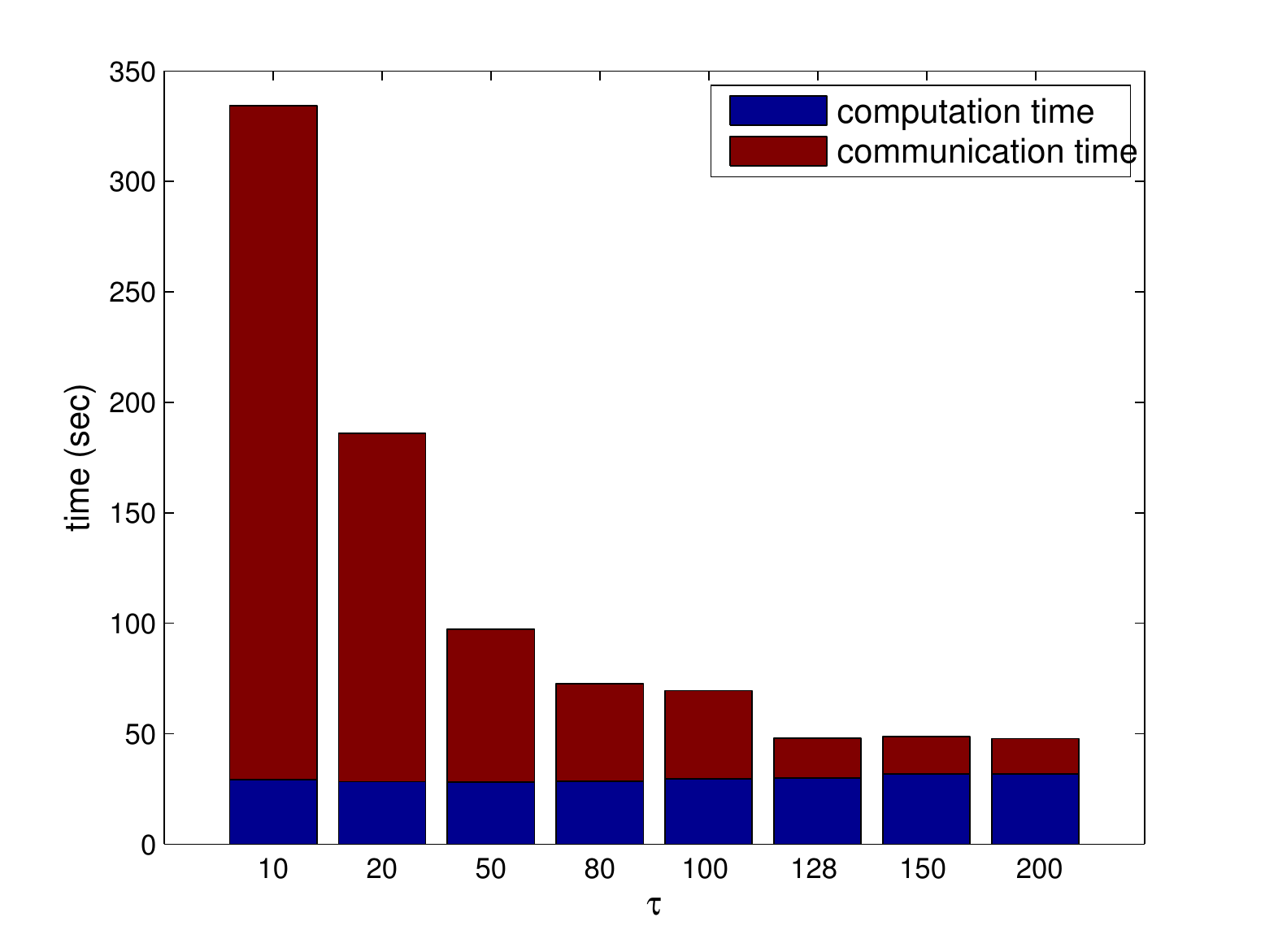}}
\vskip -.1in
\caption{Breakdown of the total time into computation time and communication time, with different 
$\tau$'s.}
\label{fig:tau1}
\end{figure}

In this experiment,  we use 128 workers.
Figure~\ref{fig:tau1} shows the time  for
\emph{distr-vr-sgd} 
to finish $mS$ tasks
(where $S=50$ and $m=\lceil N/B \rceil$)
when
$\tau$  is varied from 10 to 200.
As can be seen,
with increasing $\tau$,
higher asynchronicity is allowed,
and the communication cost is reduced significantly.

\section{Conclusion}

Existing distributed asynchronous SGD algorithms often rely on a
decaying learning rate, and 
thus suffer from a sublinear convergence rate. On the other hand, the recent delayed proximal gradient
algorithm uses a
constant learning rate and has linear convergence rate, but can only converge to within a neighborhood of the optimal solution.
In this paper, we proposed a novel distributed asynchronous SGD algorithm by integrating the merits of the stochastic 
variance reduced gradient algorithm and
delayed proximal gradient algorithm.
Using a constant learning rate, it still guarantees 
convergence to the optimal solution
at a fast linear rate.
A prototype system is implemented and run on the 
Google cloud platform. Experimental  results
show that the proposed algorithm can
reduce the communication cost significantly with the use of asynchronicity. 
Moreover, it converges much faster and yields more accurate solutions than the
state-of-the-art distributed 
asynchronous SGD 
algorithms.

\section{Acknowledgments}

This research was supported in part by
the Research Grants Council of the Hong Kong Special Administrative Region
(Grant 614012).

\bibliographystyle{aaai}
\bibliography{bibfile}

\begin{thebibliography}{}

\bibitem[\protect\citeauthoryear{Agarwal and
  Duchi}{2011}]{agarwal2012distributed}
Agarwal, A., and Duchi, J.
\newblock 2011.
\newblock Distributed delayed stochastic optimization.
\newblock In {\em Advances in Neural Information Processing Systems~24}.

\bibitem[\protect\citeauthoryear{Albrecht \bgroup et al\mbox.\egroup
  }{2006}]{albrecht2006loose}
Albrecht, J.; Tuttle, C.; Snoeren, A.; and Vahdat, A.
\newblock 2006.
\newblock Loose synchronization for large-scale networked systems.
\newblock In {\em Proceedings of the USENIX Annual Technical Conference},
  301--314.

\bibitem[\protect\citeauthoryear{Bottou}{2010}]{bottou2010large}
Bottou, L.
\newblock 2010.
\newblock Large-scale machine learning with stochastic gradient descent.
\newblock In {\em Proceedings of the International Conference on Computational
  Statistics}.
\newblock  177--186.

\bibitem[\protect\citeauthoryear{Boyd \bgroup et al\mbox.\egroup
  }{2011}]{boyd2011distributed}
Boyd, S.; Parikh, N.; Chu, E.; Peleato, B.; and Eckstein, J.
\newblock 2011.
\newblock Distributed optimization and statistical learning via the alternating
  direction method of multipliers.
\newblock {\em Foundations and Trends in Machine Learning} 3(1):1--122.

\bibitem[\protect\citeauthoryear{Dai \bgroup et al\mbox.\egroup
  }{2013}]{dai2013petuum}
Dai, W.; Wei, J.; Zheng, X.; Kim, J.~K.; Lee, S.; Yin, J.; Ho, Q.; and Xing,
  E.~P.
\newblock 2013.
\newblock Petuum: A framework for iterative-convergent distributed {ML}.
\newblock Technical Report arXiv:1312.7651.

\bibitem[\protect\citeauthoryear{Dean \bgroup et al\mbox.\egroup
  }{2012}]{dean2012large}
Dean, J.; Corrado, G.; Monga, R.; Chen, K.; Devin, M.; Mao, M.; Senior, A.;
  Tucker, P.; Yang, K.; Le, Q.~V.; et~al.
\newblock 2012.
\newblock Large scale distributed deep networks.
\newblock In {\em Advances in Neural Information Processing Systems},
  1223--1231.

\bibitem[\protect\citeauthoryear{Defazio, Bach, and
  Lacoste-Julien}{2014}]{defazio2014saga}
Defazio, A.; Bach, F.; and Lacoste-Julien, S.
\newblock 2014.
\newblock {SAGA}: A fast incremental gradient method with support for
  non-strongly convex composite objectives.
\newblock In {\em Advances in Neural Information Processing Systems},
  1646--1654.

\bibitem[\protect\citeauthoryear{Duchi, Hazan, and
  Singer}{2011}]{duchi2011adaptive}
Duchi, J.; Hazan, E.; and Singer, Y.
\newblock 2011.
\newblock Adaptive subgradient methods for online learning and stochastic
  optimization.
\newblock {\em Journal of Machine Learning Research} 12:2121--2159.

\bibitem[\protect\citeauthoryear{Feyzmahdavian, Aytekin, and
  Johansson}{2014}]{feyzmahdavian2014delayed}
Feyzmahdavian, H.~R.; Aytekin, A.; and Johansson, M.
\newblock 2014.
\newblock A delayed proximal gradient method with linear convergence rate.
\newblock In {\em Proceedings of the International Workshop on Machine Learning
  for Signal Processing},  1--6.

\bibitem[\protect\citeauthoryear{Gimpel, Das, and
  Smith}{2010}]{gimpel2010distributed}
Gimpel, K.; Das, D.; and Smith, N.~A.
\newblock 2010.
\newblock Distributed asynchronous online learning for natural language
  processing.
\newblock In {\em Proceedings of the 14th Conference on Computational Natural
  Language Learning},  213--222.

\bibitem[\protect\citeauthoryear{Ho \bgroup et al\mbox.\egroup
  }{2013}]{ho2013more}
Ho, Q.; Cipar, J.; Cui, H.; Lee, S.; Kim, J.; Gibbons, P.; Gibson, G.; Ganger,
  G.; and Xing, E.
\newblock 2013.
\newblock More effective distributed {ML} via a stale synchronous parallel
  parameter server.
\newblock In {\em Advances in Neural Information Processing Systems~26},
  1223--1231.

\bibitem[\protect\citeauthoryear{Johnson and
  Zhang}{2013}]{johnson2013accelerating}
Johnson, R., and Zhang, T.
\newblock 2013.
\newblock Accelerating stochastic gradient descent using predictive variance
  reduction.
\newblock In {\em Advances in Neural Information Processing Systems},
  315--323.

\bibitem[\protect\citeauthoryear{Li \bgroup et al\mbox.\egroup
  }{2014}]{li2014communication}
Li, M.; Andersen, D.~G.; Smola, A.~J.; and Yu, K.
\newblock 2014.
\newblock Communication efficient distributed machine learning with the
  parameter server.
\newblock In {\em Advances in Neural Information Processing Systems},  19--27.

\bibitem[\protect\citeauthoryear{Mairal}{2013}]{Mairal2013}
Mairal, J.
\newblock 2013.
\newblock Optimization with first-order surrogate functions.
\newblock In {\em Proceedings of the 30th International Conference on Machine
  Learning}.

\bibitem[\protect\citeauthoryear{Mairal}{2015}]{mairal2014incremental}
Mairal, J.
\newblock 2015.
\newblock Incremental majorization-minimization optimization with application
  to large-scale machine learning.
\newblock Technical Report arXiv:1402.4419.

\bibitem[\protect\citeauthoryear{Mania \bgroup et al\mbox.\egroup
  }{2015}]{mania2015perturbed}
Mania, H.; Pan, X.; Papailiopoulos, D.; Recht, B.; Ramchandran, K.; and Jordan,
  M.~I.
\newblock 2015.
\newblock Perturbed iterate analysis for asynchronous stochastic optimization.
\newblock Technical Report arXiv:1507.06970.

\bibitem[\protect\citeauthoryear{Nesterov}{2004}]{nesterov2004introductory}
Nesterov, Y.
\newblock 2004.
\newblock {\em Introductory Lectures on Convex Optimization}, volume~87.
\newblock Springer Science \& Business Media.

\bibitem[\protect\citeauthoryear{Niu \bgroup et al\mbox.\egroup
  }{2011}]{niu2011hogwild}
Niu, F.; Recht, B.; R{\'e}, C.; and Wright, S.
\newblock 2011.
\newblock Hogwild!: A lock-free approach to parallelizing stochastic gradient
  descent.
\newblock In {\em Advances in Neural Information Processing Systems~24}.

\bibitem[\protect\citeauthoryear{Reddi \bgroup et al\mbox.\egroup
  }{2015}]{reddi2015variance}
Reddi, S.~J.; Hefny, A.; Sra, S.; P{\'o}czos, B.; and Smola, A.
\newblock 2015.
\newblock On variance reduction in stochastic gradient descent and its
  asynchronous variants.
\newblock Technical Report arXiv:1506.06840.

\bibitem[\protect\citeauthoryear{Roux, Schmidt, and
  Bach}{2012}]{roux2012stochastic}
Roux, N.~L.; Schmidt, M.; and Bach, F.~R.
\newblock 2012.
\newblock A stochastic gradient method with an exponential convergence rate for
  finite training sets.
\newblock In {\em Advances in Neural Information Processing Systems},
  2663--2671.

\bibitem[\protect\citeauthoryear{Shalev-Shwartz and
  Zhang}{2013}]{shalev2013stochastic}
Shalev-Shwartz, S., and Zhang, T.
\newblock 2013.
\newblock Stochastic dual coordinate ascent methods for regularized loss.
\newblock {\em Journal of Machine Learning Research} 14(1):567--599.

\bibitem[\protect\citeauthoryear{Shamir, Srebro, and
  Zhang}{2014}]{shamir2014communication}
Shamir, O.; Srebro, N.; and Zhang, T.
\newblock 2014.
\newblock Communication-efficient distributed optimization using an approximate
  {N}ewton-type method.
\newblock In {\em Proceedings of the 31st International Conference on Machine
  Learning},  1000--1008.

\bibitem[\protect\citeauthoryear{Xiao and Zhang}{2014}]{xiao2014proximal}
Xiao, L., and Zhang, T.
\newblock 2014.
\newblock A proximal stochastic gradient method with progressive variance
  reduction.
\newblock {\em SIAM Journal on Optimization} 24(4):2057--2075.

\bibitem[\protect\citeauthoryear{Zhang and Kwok}{2014}]{zhang2014asynchronous}
Zhang, R., and Kwok, J.
\newblock 2014.
\newblock Asynchronous distributed {ADMM} for consensus optimization.
\newblock In {\em Proceedings of the 31st International Conference on Machine
  Learning},  1701--1709.

\end{thebibliography}
\end{document}